\documentclass[lettersize,journal]{IEEEtran}
\usepackage{amsmath,amsfonts}
\usepackage{algorithmic}
\usepackage{algorithm}
\usepackage{array}
\usepackage[caption=false,font=normalsize,labelfont=sf,textfont=sf]{subfig}
\usepackage{textcomp}
\usepackage{stfloats}
\usepackage{url}
\usepackage{verbatim}
\usepackage{graphicx}
\usepackage{cite}

\usepackage{booktabs}    
\usepackage{multirow}    
\usepackage{adjustbox}   
\usepackage{siunitx}       

\begin{document}

\title{M\textsuperscript{3}-Former: Multimodal Transformer with Mixture-of-Experts for Long-Term Vessel Trajectory Prediction}


\author{
Wenzhe Jin,
Haina Tang%
\thanks{
Wenzhe Jin, Haina Tang are with the School of Artificial Intelligence, University of Chinese Academy of Sciences, Beijing, China. (e-mail: jinwenzhe23@mails.ucas.ac.cn; hntang@ucas.ac.cn)
}
}

\markboth{Journal of \LaTeX\ Class Files,~Vol.~14, No.~8, August~2021}%
{Shell \MakeLowercase{\textit{et al.}}: A Sample Article Using IEEEtran.cls for IEEE Journals}

\IEEEpubid{0000--0000/00\$00.00~\copyright~2021 IEEE}

\maketitle

\begin{abstract}
To address the challenges of behavioral multimodality, limited semantic utilization, and long-term error accumulation in vessel trajectory prediction, this paper proposes M\textsuperscript{3}-Former, a multimodal trajectory prediction framework enhanced by large language models (LLMs). The proposed framework incorporates vessel static attributes and navigational intent as semantic priors for long-term trajectory modeling. Specifically, a unified multimodal representation space is constructed, in which static semantic information is encoded by a pre-trained LLM and aligned with dynamic trajectory features through self-attention. To jointly capture global route planning and local motion variations, a dual-granularity Mixture-of-Experts (MoE) architecture is introduced, where sequence-level experts model global navigation trends and token-level experts refine fine-grained maneuvering behaviors. In addition, a Steering-Weighted Cross-Entropy loss is designed to alleviate the long-tail distribution of sparse turning samples and improve prediction accuracy in critical maneuvering scenarios. Experiments on a real-world Danish AIS dataset demonstrate that M\textsuperscript{3}-Former consistently outperforms state-of-the-art baselines across prediction horizons from 1 to 4 hours. In the 4-hour prediction task, the proposed method reduces Average Displacement Error (ADE) and Final Displacement Error (FDE) by 4.4\% and 5.1\%, respectively, compared with the strongest baseline. Qualitative and ablation analyses further verify that semantic fusion effectively reduces long-term trajectory drift, while the dual-granularity MoE improves robustness in complex waterways and route-branching scenarios. The proposed framework establishes a semantic-guided hierarchical prediction paradigm, in which high-level navigational intent and local motion dynamics are jointly modeled for robust long-term vessel trajectory forecasting. We release our code at https://github.com/zophykim/M3former
\end{abstract}

\begin{IEEEkeywords}
Vessel Trajectory Prediction, Large Language Model, Multi-modal Fusion, Mixture of Experts, Artificial Intelligence, Maritime transportation.
\end{IEEEkeywords}

\section{Introduction}

\IEEEPARstart{W}{ith} the continuous expansion of the global maritime transportation network and the increasing intensity of shipping activities, the complexity and uncertainty of maritime traffic have significantly increased. Accurate vessel trajectory prediction is not only a fundamental problem for ensuring navigational safety, optimizing route planning, and supporting port operations, but also a key enabling technology for intelligent maritime surveillance, collision avoidance, and automated shipping scheduling~\cite{martelli2021outlook,spadon2024multi,liu2022deep}.

Compared with short-term prediction (minutes to hours), long-term trajectory prediction (hours to days) poses more fundamental modeling challenges. First, small deviations in navigation intent, environmental disturbances, or maneuvering behaviors may accumulate over time, leading to significant positional errors. Second, external constraints such as traffic lanes, port boundaries, meteorological conditions, and navigation regulations jointly shape long-term navigation strategies. More importantly, the dominant factors shift over longer time horizons: unlike short-term prediction where vessel–vessel interactions play a central role, the influence of local interactions gradually diminishes, while high-level information such as destination semantics, vessel static attributes, and navigation intent becomes the primary driver of trajectory evolution~\cite{huang2022ea,jia2023conditional,capobianco2021deep}. In addition, a single historical trajectory may correspond to multiple plausible future paths (i.e., behavioral multimodality), further increasing the difficulty of prediction.

Early studies mainly relied on physics-based and statistical models, such as Kalman filtering~\cite{simon2006optimal,perera2012maritime}, regression models, clustering-based route mining~\cite{han2021modeling,rong2020data}, and traditional machine learning methods including support vector machines and random forests~\cite{handayani2013anomaly,wang2022sequential}. Although these methods provide interpretable motion estimation in simple scenarios, their reliance on handcrafted features and simplified assumptions limits their ability to model complex maritime environments. Subsequently, recurrent neural networks such as LSTM and GRU were introduced to capture temporal dependencies in AIS sequences~\cite{chondrodima2022machine,yang2022ais,wang2020vessel}. However, recurrent models still suffer from error accumulation and insufficient modeling of multimodal future trajectories in long-term prediction.

More recently, large language models and multimodal learning have demonstrated strong capabilities in semantic modeling and cross-modal representation learning~\cite{xu2025trajectory,tang2023large,zhang2023geogpt,liu2025vtllm,nguyen2024tm}. By incorporating high-level information such as navigation intent, destination semantics, and vessel attributes into trajectory modeling, these approaches provide informative priors for long-term prediction. Meanwhile, semantic-enhanced methods~\cite{chib2025lg,chen2025semint,xu2025trajectory,ding2023mgeo,bae2024can} further explore external knowledge and intent reasoning for trajectory modeling. In addition, reformulating continuous trajectory prediction as a discrete probabilistic generation problem~\cite{nguyen2024transformer} naturally enables multimodal trajectory modeling and sampling. However, how to effectively integrate semantic information with temporal dynamics within a unified framework, while simultaneously capturing global planning and local maneuvering, remains an open challenge.
\IEEEpubidadjcol
To address these issues, we propose \mbox{M\textsuperscript{3}-Former} (Multi-modal, Multi-scale, and Mixture-of-Experts), a unified multimodal framework for long-term vessel trajectory prediction. The proposed method is built upon the principle of \emph{semantic-guided hierarchical modeling}. On the one hand, static semantic information and dynamic motion features are projected into a shared representation space and jointly modeled via attention and routing mechanisms. On the other hand, by introducing spatial discretization and probabilistic generation, the model is capable of producing multiple plausible future paths hypotheses that satisfy both physical and semantic constraints, thereby capturing the multimodal nature of vessel behavior.

From an architectural perspective, inspired by the maritime navigation paradigm of ``global planning followed by local adjustment,'' we design a dual-granularity multimodal Mixture-of-Experts (MoE) framework to achieve hierarchical modeling and collaborative fusion of global semantics and local dynamics. Specifically, at the decision level, a Per-Sequence routing mechanism is introduced, where sequence-level experts jointly model trajectory and textual modalities to capture navigation intent and global trajectory trends, providing stable macroscopic guidance for prediction. On this basis, at the perception level, a Per-Token routing mechanism is further employed, where fine-grained experts dynamically refine local temporal features to accurately model vessel maneuvers and interaction patterns. This design reduces the influence of local noise on global modeling. 
It enables the model to preserve long-term trends while remaining responsive to short-term dynamics. Moreover, the dual-granularity MoE integrates multimodal information by enforcing global consistency at the sequence level and enhancing adaptability to local variations at the token level. This complementary design consistently improves the model's ability to handle large-scale heterogeneous trajectory data and enhances robustness in long-term prediction, while keeping computational cost manageable. Combined with tailored training strategies (e.g., turn-aware weighted loss), the model further strengthens its ability to capture critical maneuvering behaviors, achieving a balance between short-term accuracy and long-term stability.

To validate the effectiveness of the proposed method, we conduct extensive experiments and ablation studies on real-world multimodal AIS datasets and compare against both traditional kinematic models and various deep learning approaches. Experimental results indicate that \mbox{M\textsuperscript{3}-Former} achieves significant improvements in both ADE and FDE, and exhibits superior physical plausibility and multimodal coverage in challenging scenarios such as turning maneuvers, route bifurcations, and complex port environments.

The main contributions of this paper are summarized as follows:
\begin{itemize}
    \item We formulate long-term vessel trajectory prediction as a semantic-guided forecasting problem by incorporating static vessel attributes as navigational priors.
    \item We design a dual-granularity multimodal Mixture-of-Experts architecture following a global-to-local paradigm. Through Per-Sequence and Per-Token routing mechanisms, the model achieves joint optimization of global navigation strategy modeling and fine-grained cross-modal feature alignment.
    \item We conduct comprehensive evaluations on real-world multimodal maritime datasets. Experimental results indicate that the proposed method outperforms existing approaches in prediction accuracy, long-term consistency, and robustness under complex environments.
\end{itemize}

\section{Problem Definition}\label{sec:4.2}

Vessel trajectories exhibit inherent intention diversity, i.e., vessels departing from the same origin may head toward different destinations. In this work, we focus on long-term vessel trajectory prediction based on a large-model-driven multimodal fusion framework, aiming to support full-route planning and operational management over horizons ranging from hours to days. The goal is to effectively capture spatio-temporal dependencies in trajectories as well as external influencing factors.

We assume that the input to the trajectory prediction system consists of both dynamic and static information. The dynamic trajectory of the $i$-th vessel is defined as

\begin{equation}
\begin{aligned}
y^i = \{ p_t^i = (&\text{lat}_t^i, \text{lon}_t^i, \text{sog}_t^i, \text{cog}_t^i) \mid t=1,\dots,T \}, \\
& \forall i \in \{1,2,\dots,N\}
\end{aligned}
\end{equation}

where $\text{sog}$ denotes Speed Over Ground (SOG), and $\text{cog}$ denotes Course Over Ground (COG). Each $p_t^i$ consists of latitude, longitude, speed over ground, and course over ground, and $T$ represents the trajectory length. The static features are defined as
\begin{equation}
\begin{aligned}
s^i = (&\text{Destination}^i, \text{ShipType}^i, \text{Width}^i, \text{Length}^i), \\ 
&  \forall i \in \{1,2,\dots,N\},
\end{aligned}
\end{equation}

which are obtained from AIS data, including destination, vessel type, width, and length.

The objective of long-term trajectory prediction is to learn the conditional distribution
\begin{equation}
p(y_t^i \mid y_{1:t-1}^i, s^i), \quad \forall i \in \{1,2,\dots,N\},
\end{equation}
so as to predict future trajectory points given historical trajectories and static information. This formulation enables the integration of dynamic motion patterns and static intent, thereby improving the accuracy of long-term route prediction.

\section{Methodology}\label{sec:4.3}

The overall design follows a semantic-guided hierarchical modeling principle, where semantic priors determine global trajectory evolution and dynamic motion patterns refine local behaviors.

\begin{figure*}[!t]  
\centering
\includegraphics[width=\textwidth]{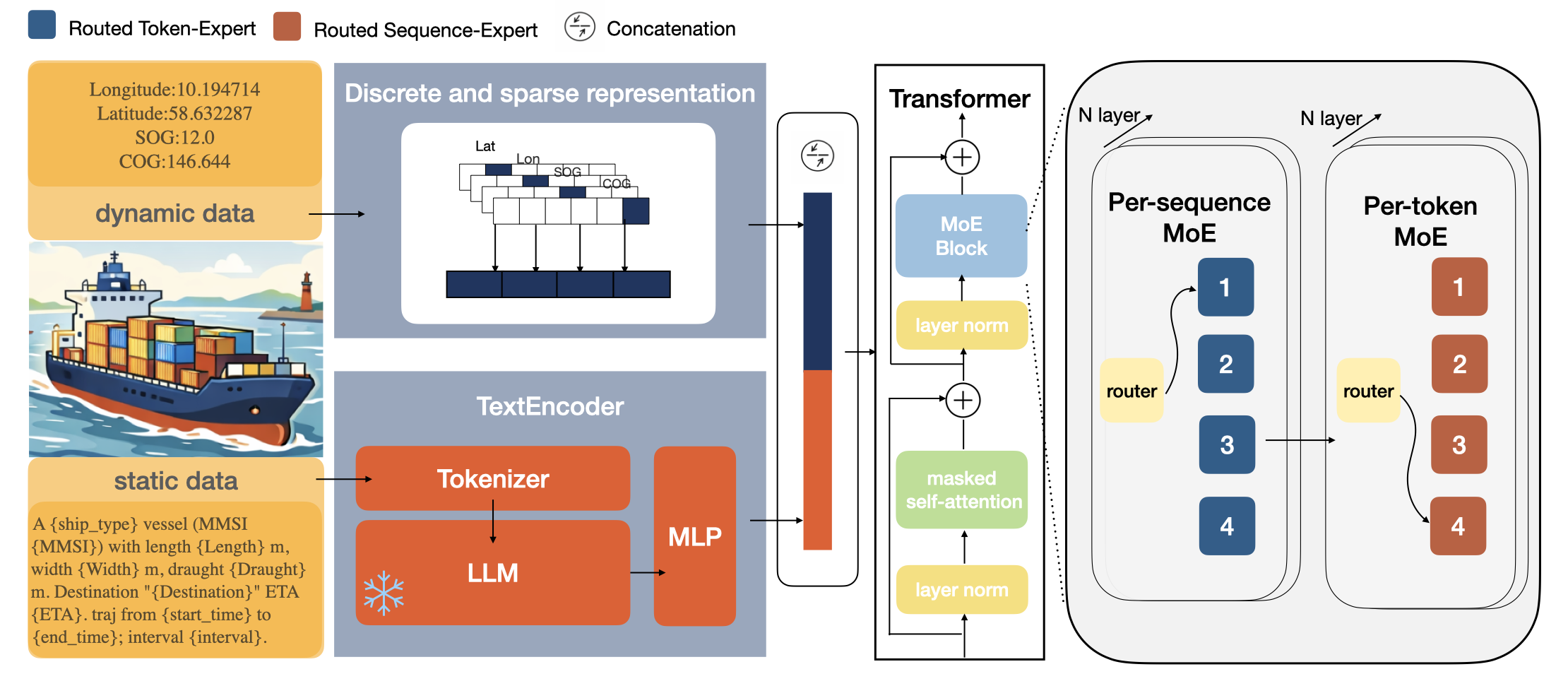}  
\caption{Overall Architecture of the Proposed M\textsuperscript{3}-Former.}
\label{fig:model}
\end{figure*}

\subsection{Multimodal Feature Fusion via Large Language Models}

Long-term vessel trajectory prediction plays a crucial role in maritime supervision, route planning, and navigation safety. However, long-term prediction faces significant challenges due to behavioral multimodality, where the same historical state may evolve into multiple potential future endpoints. This uncertainty arises from the interaction of local geographic constraints (e.g., shipping lanes, reefs) and global navigational intent (e.g., vessel specifications, destination, estimated time of arrival, ETA). To address the limitation of existing approaches that overly rely on dynamic features while neglecting semantic guidance, we introduce the semantic representation capability of large language models (LLMs) and construct a unified multimodal feature fusion framework based on self-attention to jointly model trajectory dynamics and high-level semantic information.

We formulate the trajectory prediction task as a conditional probability learning problem. For a target vessel $i$, its input feature vector $\mathbf{x}_i$ consists of dynamic motion features $\mathbf{d}_{i,t}$ (including latitude, longitude, SOG, COG, etc.) and static semantic features $\mathbf{s}_i$. As shown in Fig.~\ref{fig:model}, a concrete input example is presented.

\begin{enumerate}
    \item \textbf{Unified Representation of Heterogeneous Features:} Inspired by multimodal LLMs, we project heterogeneous information into a unified latent space. Specifically, structured static information is first converted into a natural language description (e.g., ``A Container vessel (MMSI 477307200) with length 366.0 m, width 48.2 m, draught 13.5 m. Destination `CNTAO' ETA 2023-07-15 14:00. traj from 2023-07-10 08:30:00 to 2023-07-10 16:30:00; interval 10 min.''), which is then encoded by a frozen pre-trained LLM (e.g., LLaMA) to obtain semantic embeddings $\mathbf{E}_{\text{text}}$. To adapt to downstream tasks, these embeddings are further mapped and refined via a nonlinear projection layer. Dynamic trajectory features are linearly mapped to obtain temporal embeddings $\mathbf{E}_{\text{dyn}}$. During fusion, the text embeddings are prepended as prefix tokens and concatenated with trajectory tokens along the sequence dimension. The combined sequence is then fed into a shared Transformer encoder, where self-attention enables cross-modal interaction and alignment. The resulting fused multimodal representation can be formalized as:
    \begin{equation}
        \mathbf{E}_{\text{fused}} = \text{LayerNorm}\Big(\text{Linear}\big([\mathbf{E}_{\text{dyn}} ; \mathbf{E}_{\text{text}}]\big)\Big),
    \end{equation}
    where $[\cdot ; \cdot]$ denotes feature concatenation. This design allows the model to automatically learn the modulation effect of semantic information on trajectory evolution in the self-attention space. We compared multiple fusion strategies, including weighted fusion, gating mechanisms, and cross-attention, and found that unified self-attention modeling achieves the best performance in prediction accuracy and multimodal distribution modeling, consistent with cross-modal alignment results in prior multimodal methods such as ViLBERT~\cite{lu2019vilbert} and CLIP~\cite{radford2021learning}. Implementation-wise, the process corresponds to first encoding static information with an LLM to obtain token-level representations, then concatenating them with trajectory embeddings along the sequence dimension for main model processing.

    \item \textbf{Spatial Discretization and Probabilistic Modeling:} To enhance the model's ability to capture multimodal distributions, we follow the design of TrAISformer~\cite{li2024traisformer} and convert the continuous coordinate regression task into a discrete spatial classification task. The study area is divided into spatial bins, projecting continuous coordinates to discrete tokens in a set $\mathcal{V}$. At each time step $t$, the model outputs logits over the discrete positions and explicitly models the conditional probability as:
    \begin{equation}
        P(y_{t+1} \mid y_{1:t}, \mathbf{s}_i) = \text{Softmax}\big(f_{\theta}(\mathbf{E}_{\text{fused}})\big).
    \end{equation}
\end{enumerate}

This mechanism allows the model to generate multiple candidate trajectories that respect physical constraints and semantic logic, effectively capturing the uncertainty and diversity of potential navigational intents in long-term prediction scenarios.

\subsection{Dual-Granularity Mixture-of-Experts (MoE) Model}

To capture trajectory evolution patterns across different temporal scales and semantic hierarchies, we introduce a dual-granularity Mixture-of-Experts (MoE) architecture within the Transformer backbone. This design implements decoupled expert routing mechanisms at both the sequence and token levels, enabling hierarchical modeling of global navigational intent and local motion patterns. By integrating multimodal information (trajectory features and static vessel attributes), the model achieves enhanced expressiveness for complex trajectory distributions.

\begin{enumerate}

\item \textbf{Sequence-Level MoE (Per-Sequence MoE):} Sequence-level MoE focuses on global navigational intent and long-term behavior patterns. The historical sequence is aggregated to obtain a global context vector:
\begin{equation}
    \mathbf{z} = \text{Pool}(\{\mathbf{h}_1, \mathbf{h}_2, \dots, \mathbf{h}_T\}),
\end{equation}
where $\text{Pool}(\cdot)$ denotes average pooling.

Based on the global context, a gating network generates sequence-level expert weights:
\begin{equation}
    \mathbf{g}^{\text{seq}} = \text{Softmax}(\mathbf{W}_s \mathbf{z}),
\end{equation}
with $\mathbf{W}_s \in \mathbb{R}^{k \times d}$ and $k$ being the number of sequence-level experts.

The sequence-level MoE output representation is computed as:
\begin{equation}
    \mathbf{h}^{\text{seq}} = \sum_{j=1}^{k} g^{\text{seq}}_{j} \cdot E^{\text{seq}}_j(\mathbf{z}),
\end{equation}
providing long-term navigational constraints such as destination-driven route selection, low-speed maneuvering in ports or channels, and cruising behavior in open waters. Multimodal information is naturally integrated at this stage, allowing experts to learn globally coherent navigation strategies conditioned on vessel attributes and operational constraints.

\item \textbf{Token-Level MoE (Per-Token MoE):} Token-level MoE operates on each time-step representation $\mathbf{h}_t \in \mathbb{R}^{d}$ to model local motion dynamics and fine-grained multimodal fusion:
\begin{equation}
    \mathbf{g}_t = \text{Softmax}(\mathbf{W}_g \mathbf{h}_t),
\end{equation}
where $\mathbf{W}_g \in \mathbb{R}^{k \times d}$, $k$ is the number of experts, and $\mathbf{g}_t$ denotes the activation weights for each expert.

The token-level MoE output is given by:
\begin{equation}
    \mathbf{h}_t^{\text{moe}} = \sum_{j=1}^{k} g_{t,j} \cdot E_j(\mathbf{h}_t),
\end{equation}
enabling the model to dynamically select the most suitable expert based on the current motion state (e.g., speed or course changes). Different experts can specialize in local patterns such as straight-line cruising, gradual corrections, or sharp turns, and adaptively weight multimodal information to enhance the representation of fine-grained motion behavior.

\end{enumerate}

Experimental analysis (see Sec.~\ref{sec:4.4.7.1}) indicates that the MoE-S$\rightarrow$T configuration—applying sequence-level MoE prior to token-level MoE—outperforms the MoE-T$\rightarrow$S alternative. This ordering aligns with navigational behavior logic: vessels typically plan overall routes before adjusting heading and speed based on local dynamics. Extracting global trends first, followed by fine-grained adjustments, facilitates the integration of global and local information while minimizing interference from local noise.

Overall, the dual-granularity MoE achieves coordinated modeling of short-term dynamic changes and long-term navigational intent through hierarchical local-global modeling and multimodal fusion. The token-level MoE emphasizes responsiveness to local motion patterns, while the sequence-level MoE provides stable global constraints, together enhancing the model’s ability to handle large-scale, heterogeneous trajectory data with robust long-term predictive performance.

\subsection{Steering-Weighted Auxiliary Loss Design}\label{sec:loss function}

To address the long-tailed distribution of vessel trajectory data, where straight-line sailing dominates while turning behaviors are relatively sparse, we introduce a Steering-Weighted Cross-Entropy (SWCE) loss. Conventional training objectives tend to be dominated by frequent straight-motion samples, which often leads to delayed responses or over-smoothed predictions near critical maneuvering regions.

Since the proposed framework formulates trajectory prediction as a discrete token generation task, the model independently predicts four variables at each time step: latitude, longitude, speed over ground (SOG), and course over ground (COG). The basic prediction loss is defined as the sum of the cross-entropy losses of these four variables:
\begin{equation}
\mathcal{L}_{\text{base}}
=
\mathcal{L}_{\text{lat}}
+
\mathcal{L}_{\text{lon}}
+
\mathcal{L}_{\text{sog}}
+
\mathcal{L}_{\text{cog}},
\end{equation}
where each term denotes the categorical cross-entropy loss of the corresponding discretized variable.

To emphasize rare turning samples, the heading change between consecutive time steps is first computed as
\begin{equation}
\Delta \mathrm{COG}_{i,t}
=
\left|
\mathrm{COG}_{i,t}
-
\mathrm{COG}_{i,t-1}
\right|.
\end{equation}

A steering-aware weighting factor is then defined as
\begin{equation}
\lambda_{i,t}
=
1+\alpha \cdot \mathbf{1}
\left(
\Delta \mathrm{COG}_{i,t}>\tau
\right),
\end{equation}
where $\mathbf{1}(\cdot)$ is the indicator function, $\tau$ is the turning threshold, and $\alpha$ controls the additional penalty assigned to turning samples.

The final weighted loss is therefore written as
\begin{equation}
\mathcal{L}
=
\frac{1}{N}
\sum_{i=1}^{N}
\sum_{t=1}^{T}
\lambda_{i,t}
\,
\mathcal{L}_{\text{base}}^{(i,t)}.
\end{equation}

By increasing the contribution of turning samples during optimization, the proposed loss encourages the model to better capture critical maneuvering behaviors. This mechanism reduces the bias toward dominant straight-line trajectories and improves prediction accuracy in complex navigation scenarios such as channel transitions, route bifurcations, and collision-avoidance maneuvers.

\section{Experiments and Analysis}\label{sec:4.4}

\subsection{Dataset and Preprocessing}\label{sec:4.4.1}

\begin{figure}[!t]
\centering
\includegraphics[width=2.5in]{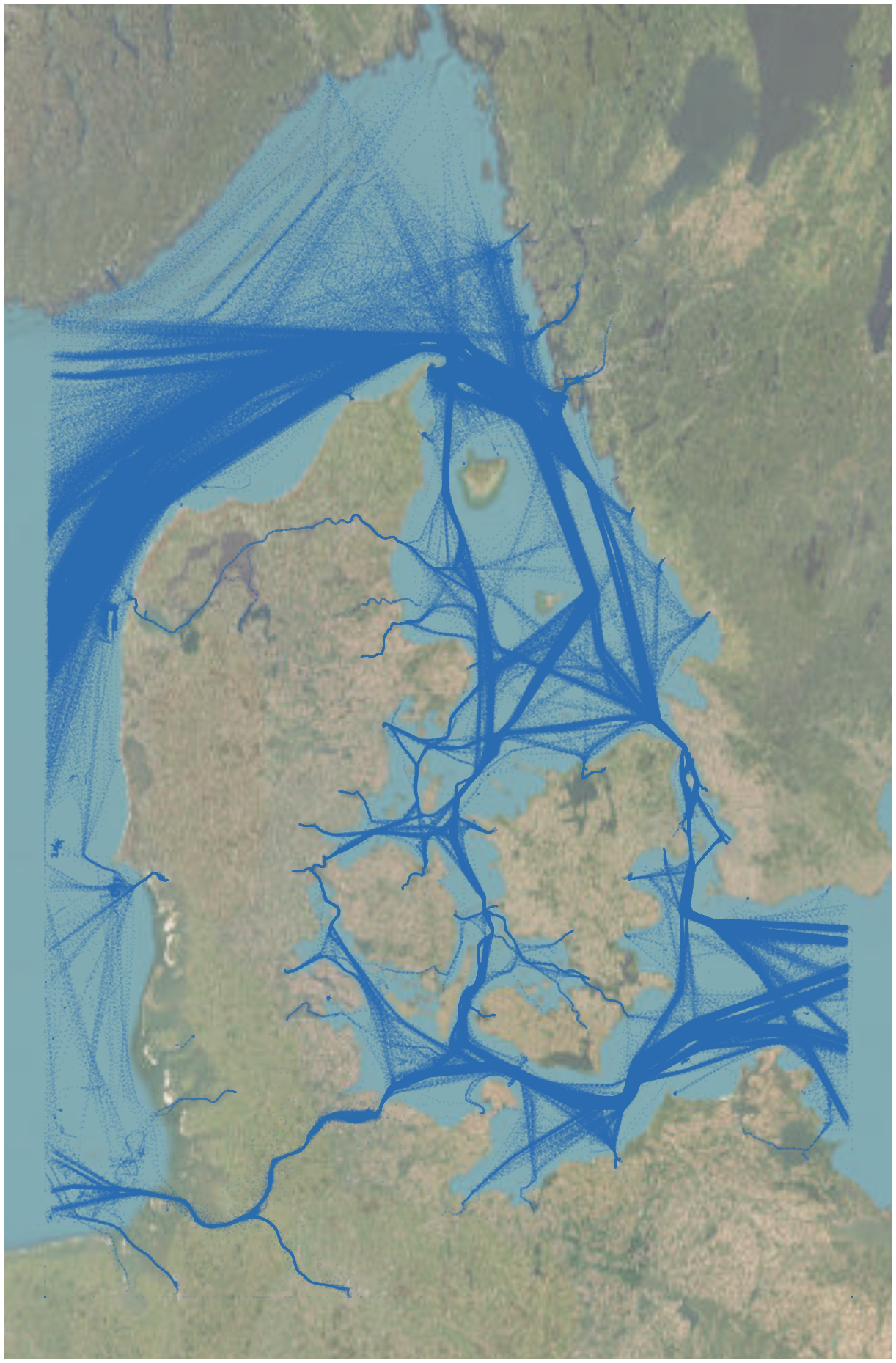}
\caption{Geographical Distribution of AIS Data in Danish Waters.}
\label{fig:dataset}
\end{figure}

For this study on long-term trajectory prediction via multimodal fusion, we utilize a publicly available real-world AIS dataset provided by the Danish Maritime Authority (DMA), as illustrated in Fig.~\ref{fig:dataset}. The dataset covers vessel navigation records from January 1, 2023, to March 31, 2023, featuring high spatiotemporal continuity and strong representativeness of real-world maritime operations. This provides a rich basis for modeling complex vessel behavior patterns. To fully exploit vessel motion patterns, the raw AIS data is abstracted and reconstructed into two categories of features: dynamic motion features and static attribute features.

\textbf{Dynamic Motion Features:} These primarily consist of time-varying sequential information characterizing vessel motion states. Specifically, they include spatial trajectories formed by longitude and latitude, as well as Speed Over Ground (SOG) and Course Over Ground (COG). Together, these features describe the vessel’s kinematic attributes and form the core basis for modeling both short-term motion trends and long-term trajectory evolution.

\textbf{Static Attribute Features:} These enhance the model’s ability to represent behavior patterns of different vessel types. They include the vessel’s unique MMSI identifier, physical attributes such as length, width, and draught, and vessel type. Additionally, semantic information such as destination and estimated time of arrival (ETA) is incorporated to capture navigational intent, providing endpoint constraints and high-level guidance for long-term trajectory prediction.

\begin{figure}[t]
\centering
\small

\fbox{
\begin{minipage}{0.9\linewidth}
\textbf{Algorithm 1: AIS Trajectory Preprocessing Pipeline}

\rule{\linewidth}{0.5pt}

\textbf{Input:} Raw AIS CSV dataset $\mathcal{D}$, time interval $\Delta t$ \\
\textbf{Output:} Interpolated trajectory set $\mathcal{T}$

\vspace{0.3em}
\rule{\linewidth}{0.5pt}

\textbf{For each} file $d \in \mathcal{D}$ \textbf{do}
\begin{enumerate}
  \item Load CSV file and convert timestamps to Unix time
  \item Filter invalid MMSI entries, missing values, and abnormal records
  \item Select cargo ships and tankers under valid navigation status
  \item Apply spatial and speed constraints
\end{enumerate}

\textbf{For each} MMSI $m$ \textbf{do}
\begin{enumerate}
  \item Extract trajectory $\tau_m$ and sort by timestamp
  \item Remove duplicated timestamps
\end{enumerate}

\textbf{For each} trajectory $\tau$ \textbf{do}
\begin{enumerate}
  \item \textbf{If} $\max(\Delta t_i) > T_{\max}$ \textbf{or}
        total duration $< T_{\min}$ \textbf{or}
        trajectory length $< N_{\min}$ \textbf{then discard}
  \item Remove overly dense points (temporal filtering)
  \item Align timestamps to a uniform grid with interval $\Delta t$
  \item Interpolation:
    \begin{itemize}
      \item Longitude/Latitude $\rightarrow$ Cubic spline interpolation
      \item Speed Over Ground (SOG) $\rightarrow$ Piecewise Cubic Hermite Interpolation (PCHIP)
      \item Course Over Ground (COG) / Heading $\rightarrow$ Circular interpolation
    \end{itemize}
  \item Attach vessel static attributes
  \item Store interpolated trajectory
\end{enumerate}

\rule{\linewidth}{0.5pt}

\textbf{return} $\mathcal{T}$

\end{minipage}
}
\caption{Pseudo-code framework of AIS trajectory preprocessing pipeline.}
\label{fig:preprocess_pipeline}

\end{figure}

Considering the noise, irregular sampling, and missing signals commonly present in AIS data collection and transmission, we design a systematic preprocessing pipeline (see Fig.~\ref{fig:preprocess_pipeline}) to convert raw data into regularized, model-ready trajectory sequences.

After preprocessing, the training set contains 1,923,380 valid samples covering 4,089 vessels, resulting in 26,374 trajectory segments and 4,407 unique destinations. Cargo vessels and tankers contribute 1,345,435 and 577,945 samples, respectively. The average trajectory length is 72.93 time steps, with a maximum of 216 and a minimum of 22. The resulting dataset effectively preserves dynamic motion patterns and static semantic attributes of vessels in real maritime environments, providing a reliable foundation for long-term multimodal trajectory prediction.

\subsection{Implementation Detail}\label{sec:4.4.2}

All experiments are conducted on a server equipped with four NVIDIA A100-SXM4-80GB GPUs and an Intel Xeon 128-core CPU. The implementation is based on PyTorch with distributed training support.

On the Danish Waters AIS dataset, the model is trained with an input sequence length of 36 and a prediction horizon of 24 time steps. Using a batch size of 256, training for 5 epochs takes approximately 6 hours on four GPUs.

During inference, the model performs auto-regressive trajectory generation with a batch size of 400. The average inference time for the full test set is about 14 minutes, demonstrating that the proposed method maintains both computational efficiency and scalability.

\subsection{Hyperparameter Settings}\label{sec:4.4.3}

To ensure reproducibility, all experiments are conducted with a fixed random seed of 2026. The input sequence length is set to $T=36$, and the prediction horizon is $T'=24$.

The model uses dynamic features including longitude, latitude, speed over ground (SOG), and course over ground (COG), along with static attributes such as vessel type, dimensions, destination, and ETA.

In M\textsuperscript{3}-Former, trajectory features are projected into a 768-dimensional embedding space using separate embedding layers. The Transformer backbone adopts a GPT-style causal self-attention architecture with 8 layers and 8 attention heads.

The Mixture-of-Experts module consists of 4 experts with Top-1 routing and a load balancing weight of $1\times10^{-4}$. The model is trained using the Adam optimizer with a learning rate of 1e-4, and automatic mixed precision (AMP) is optionally applied.

Training is performed for up to 10 epochs with early stopping (patience=5). Each experiment is repeated 10 times, and the best model based on validation loss is selected for evaluation.

\subsection{Evaluation Metrics}\label{sec:4.4.4}

To comprehensively assess the performance of the proposed model in vessel trajectory prediction, we adopt the Average Displacement Error (ADE) and Final Displacement Error (FDE) as the primary evaluation metrics. ADE measures the average spatial deviation of the predicted trajectory from the ground truth over the entire time sequence, while FDE evaluates the distance error between the predicted and true final positions. These metrics jointly reflect the prediction accuracy from both the overall trajectory shape and endpoint perspectives. Formally, for a trajectory of length $T$, they are defined as:

\begin{equation}\label{eq:ade}
    \text{ADE} = \frac{1}{T} \sum_{t=1}^{T} \| \hat{\mathbf{p}}_t - \mathbf{p}_t \|_2
\end{equation}

\begin{equation}\label{eq:fde}
    \text{FDE} = \| \hat{\mathbf{p}}_T - \mathbf{p}_T \|_2
\end{equation}

where $\hat{\mathbf{p}}_t$ and $\mathbf{p}_t$ denote the predicted and ground-truth positions at time step $t$, respectively, and $\|\cdot\|_2$ is the Euclidean distance.

During training, the model is optimized using the steering-weighted multi-task cross-entropy loss defined in Section \ref{sec:loss function}. During inference, the predicted discrete tokens are transformed back into continuous geographic coordinates for ADE and FDE evaluation.

For testing, historical trajectory sequences are provided as model inputs, and future ground-truth trajectories serve as labels for performance evaluation. All metrics are computed over the entire test set and averaged to report the overall prediction performance. To reduce the influence of randomness, each experiment is repeated multiple times, and the mean of these runs is reported as the final result.

\subsection{Quantitative Evaluation}\label{sec:4.4.5}

\begin{table*}[!t]
\caption{Quantitative Comparison with Baseline Models under Different Prediction Horizons}\label{tab:model_compare}
\centering
\small  
\begin{tabular}{l c c c c c c c c}
\toprule
\multirow{2}{*}{Model} 
& \multicolumn{2}{c}{1h (S6)} 
& \multicolumn{2}{c}{2h (S12)} 
& \multicolumn{2}{c}{3h (S18)} 
& \multicolumn{2}{c}{4h (S24)} \\
\cmidrule(lr){2-3} \cmidrule(lr){4-5} \cmidrule(lr){6-7} \cmidrule(l){8-9}
& ADE$\downarrow$ & FDE$\downarrow$
& ADE$\downarrow$ & FDE$\downarrow$
& ADE$\downarrow$ & FDE$\downarrow$
& ADE$\downarrow$ & FDE$\downarrow$ \\
\midrule
CV & 3.9717 & 7.1912 & 8.3058 & 16.8313 & 13.3173 & 28.0776 & 18.7701 & 40.1790 \\
KF~\cite{simon2006optimal} & 4.0263 & 7.1368 & 8.2412 & 16.6029 & 13.1742 & 27.7648 & 18.5298 & 39.6351 \\
Seq2Seq~\cite{zhao2025ship} & 4.2661 & 5.8374 & 6.3577 & 10.2458 & 8.4571 & 14.4143 & 10.6793 & 19.6058 \\
TCN~\cite{farahnakian2025maritime} & 4.0252 & 5.4136 & 5.6332 & 8.6152 & 7.4096 & 12.8933 & 9.4732 & 17.7865 \\
GRU~\cite{wang2020vessel} & 3.8014 & 5.1171 & 5.4723 & 8.7089 & 7.2802 & 12.5637 & 9.2527 & 17.3268 \\
TrAISformer~\cite{li2024traisformer} & \underline{1.5768} & \underline{2.7409} 
& \underline{3.1175} & \underline{6.1124} 
& \underline{4.8694} & \underline{10.0101} 
& \underline{6.7732} & \underline{14.3445} \\
\textbf{M\textsuperscript{3}-Former} & \textbf{1.5537} & \textbf{2.6763} 
& \textbf{3.0280} & \textbf{5.8744} 
& \textbf{4.6790} & \textbf{9.5135} 
& \textbf{6.4727} & \textbf{13.6142} \\
\bottomrule
\end{tabular}
\end{table*}

To systematically evaluate the performance of the proposed model in long-term vessel trajectory prediction, we compare M\textsuperscript{3}-Former with representative baselines ranging from traditional motion models to recent deep learning methods.

\begin{enumerate}
\item \textbf{CV}: A Constant Velocity model that extrapolates future vessel positions by assuming unchanged speed and heading. This method serves as a simple deterministic baseline for stable motion scenarios.

\item \textbf{KF}~\cite{simon2006optimal}: A Kalman Filter based model that recursively estimates vessel states from noisy observations under linear dynamic assumptions. It provides a stronger statistical baseline than CV for short-term prediction.

\item \textbf{Seq2Seq}~\cite{zhao2025ship}: An encoder--decoder recurrent network that learns temporal dependencies from historical trajectories and generates future positions sequentially.

\item \textbf{GRU}~\cite{wang2020vessel}: A gated recurrent neural network that improves long-sequence modeling by mitigating vanishing gradients and has been widely used in vessel trajectory prediction.

\item \textbf{TCN}~\cite{farahnakian2025maritime}: A Temporal Convolutional Network that captures long-range dependencies using causal dilated convolutions while enabling efficient parallel computation.

\item \textbf{TrAISformer}~\cite{li2024traisformer}: A Transformer-based model that discretizes continuous trajectories into spatial tokens and models long-range spatiotemporal dependencies through self-attention.
\end{enumerate}

The experimental results, summarized in Table~\ref{tab:model_compare}, employ ADE and FDE as metrics. While CV and KF provide basic references for short-term tasks, their errors escalate rapidly over longer horizons; for a 4-hour prediction, their FDE exceeds 39, significantly higher than deep learning counterparts. In contrast, GRU achieves ADE = 9.2527 and FDE = 17.3268, demonstrating the advantage of non-linear sequence modeling.

TrAISformer further reduces errors, achieving ADE = 6.7732 and FDE = 14.3445 in the 4-hour task, outperforming GRU by 26.8\% and 17.2\% respectively. This confirms the efficacy of discretized spatiotemporal modeling. Building upon these, the proposed M\textsuperscript{3}-Former incorporates semantic features from a large language model and utilizes a dual-granularity Mixture-of-Experts (MoE) structure. Consequently, it achieves state-of-the-art results across all horizons. For the 4-hour prediction, M\textsuperscript{3}-Former reaches ADE = 6.4727 and FDE = 13.6142, improving over TrAISformer by 4.4\% and 5.1\%. Notably, M\textsuperscript{3}-Former exhibits a significantly slower error growth rate as the horizon extends (e.g., ADE increasing from 1.5537 to 6.4727 compared to GRU's growth from 3.8014 to 9.2527), indicating that multimodal semantic priors effectively mitigate long-term drift.

In summary, the results validate that integrating high-level vessel intent via LLMs and capturing complex patterns through the dual-granularity MoE structure significantly enhances both the accuracy and stability of long-term vessel trajectory prediction.

\subsection{Qualitative Analysis}\label{sec:4.4.6}

\subsubsection{Prediction Comparisons in Representative Scenarios}

\begin{figure*}[!t]  
\centering
\includegraphics[width=\textwidth]{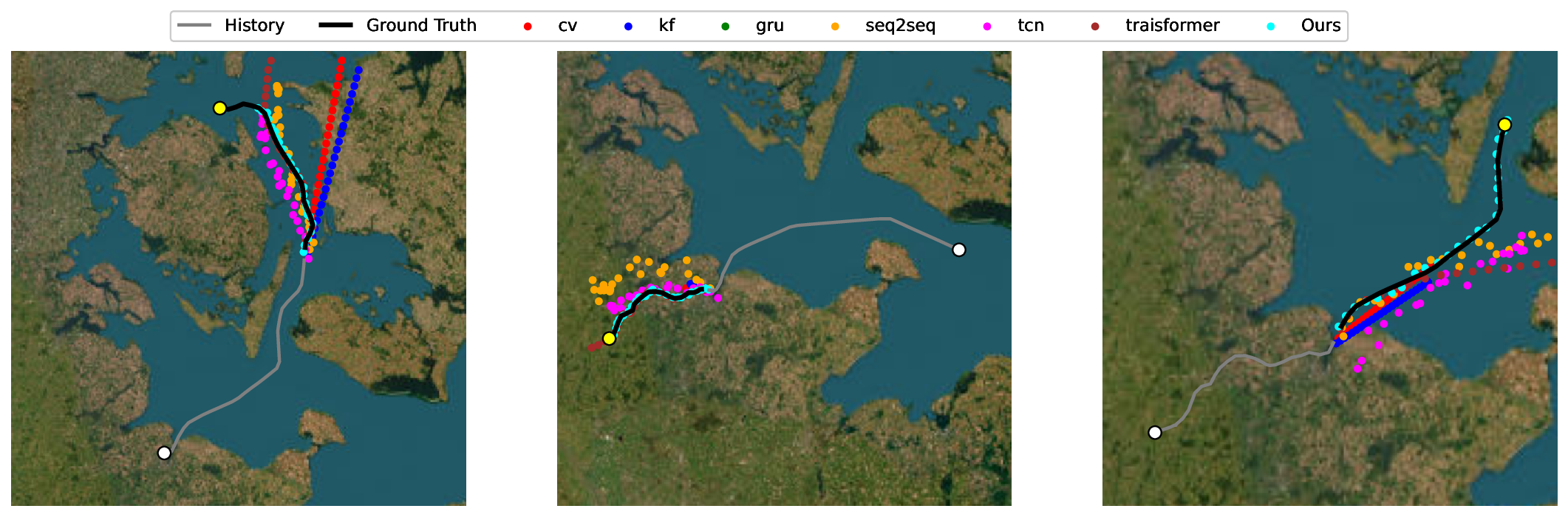}  
\caption{Qualitative Comparison of Long-Term Trajectory Prediction Results.}
\label{fig:comparison}
\end{figure*}

To visually assess the effectiveness of the proposed multimodal vessel trajectory prediction model (M\textsuperscript{3}-Former), Fig.~\ref{fig:comparison} presents three representative navigation scenarios selected from the test set. The results indicate that the key observations in each subplot, analyze the underlying model behaviors, and highlight the advantages of our method in terms of physical plausibility and robustness in real maritime conditions.

In the left subplot, representing a sharp-turn scenario, traditional kinematic models (CV, KF) fail to capture short-term nonlinear maneuvering intentions due to their constant-speed or constant-acceleration assumptions, resulting in straight-line drift. Sequence-based models (RNNs or convolutional) exhibit initial lag in turn prediction—i.e., the predicted trajectory lags behind the Ground Truth during directional changes.  
The advantage of our method lies in the MoE gating network, which dynamically allocates expert weights based on high-order dynamic features such as current speed and heading change rate. This quickly activates sub-models specialized in high-curvature motion. Consequently, the cyan curve (M\textsuperscript{3}-Former) closely follows the Ground Truth from the onset of the turn, significantly reducing lag errors and lateral deviation.

The middle subplot illustrates a continuous heading-change scenario in narrow waters or near complex shorelines. Such environments pose two major challenges: (1) terrain constraints (shorelines, shallow regions) limit trajectory feasibility, and (2) vessel behavior is highly constrained, potentially influenced by navigation aids or implicit rules. Models relying solely on historical coordinates are prone to “land crossing” or generate physically implausible oscillations.  
Our approach employs multimodal fusion to jointly encode AIS static attributes (e.g., vessel type, draught, destination) with dynamic trajectory features, endowing the model with environmental awareness. Additionally, motion-kinematics constraints are incorporated into the loss function to penalize non-compliant heading change rates. As a result, M\textsuperscript{3}-Former maintains smooth, spatially consistent predictions even in regions with high curvature, avoiding land crossing or unrealistic maneuvers.

The right subplot depicts a typical bifurcating channel scenario, where multiple plausible trajectories exist from the same historical starting point (i.e., multimodal behavior). Models lacking intent information (trajectory-only) often produce “mean predictions” or erroneously concentrate on a single path. In contrast, our model leverages AIS static fields and the MoE-based multi-expert structure, allowing different experts to capture distinct potential headings or intentions. The gating network assigns probabilistic weights based on current states (e.g., speed profile, docking tendency, vessel type), effectively modeling the multimodal distribution at branch points. Experiments show that while TrAISformer outperforms simpler baselines, it still struggles to consistently select the correct branch in the absence of intent information. M\textsuperscript{3}-Former, however, reliably predicts trajectories aligned with the actual branch, due to the synergistic effect of expert gating and intent representation.

Overall, baseline models (CV, KF) extend trajectories along linear physical assumptions, leading to straight-line drift in high-maneuver scenarios. Sequence-based models (GRU, Seq2Seq, TCN) can capture short-term motion trends but often produce lag, land crossing, or diverging predictions when handling sharp turns, near-shore navigation, or channel bifurcations. In contrast, our method integrates multimodal information (AIS static attributes + dynamic trajectory features) and employs a Mixture-of-Experts (MoE) architecture with targeted kinematics constraints, consistently yielding predictions closer to the true trajectories across diverse and challenging scenarios.

\begin{table}[!t]
\centering
\caption{Effect of Destination Information on Long-Term Prediction Performance}\label{tab:destination-ablation}
\small
\begin{tabular}{@{}l c c c c@{}}
\toprule
\multirow{2}{*}{Prediction Horizon} 
& \multicolumn{2}{c}{M\textsuperscript{3}-Former} 
& \multicolumn{2}{c}{Without Destination} \\
\cmidrule(lr){2-3} \cmidrule(l){4-5}
& ADE$\downarrow$ & FDE$\downarrow$ 
& ADE$\downarrow$ & FDE$\downarrow$ \\
\midrule
6 steps  & 1.5537 & 2.6763 & 1.5684 & 2.7050 \\
12 steps & 3.0280 & 5.8744 & 3.0706 & 5.9958 \\
18 steps & 4.6790 & 9.5135 & 4.7824 & 9.8096 \\
24 steps & 6.4727 & 13.6142 & 6.6444 & 14.0300 \\
\bottomrule
\end{tabular}
\end{table}

\subsubsection{Ablation Study on Destination Information}\label{sec:4.4.7}

To further investigate the impact of the AIS textual field \emph{Destination} on model performance, we conduct an ablation study by replacing the original destination information with an \texttt{UNK} token and comparing the results with the full model, as shown in Table~\ref{tab:destination-ablation}.

Overall, removing destination information consistently increases both ADE and FDE across all prediction horizons (Steps 6--24), with errors gradually amplifying as the prediction horizon extends. This indicates that destination information plays a sustained and cumulative role in long-term trajectory forecasting.

Specifically, for short-term predictions (Step 6), the performance gap is relatively minor (ADE increases by only ~0.015 NM), suggesting that the model primarily relies on historical motion states such as speed and heading for very short-term forecasts. However, as the prediction horizon extends, the discrepancy grows: at Step 24, ADE and FDE increase by approximately 0.17 NM and 0.42 NM, respectively, demonstrating a significant accumulation of error. This trend highlights that in long-term prediction, the absence of destination constraints leads the model to gradually deviate from the true navigation intent, resulting in cumulative error growth.

Mechanistically, destination information provides a form of global intention constraint. In scenarios with channel bifurcations or multimodal paths, this information effectively reduces the feasible trajectory space and guides the model toward the correct route. In its absence, the model relies solely on local motion patterns for extrapolation, which often produces “averaged” predictions or mis-selected paths in branching areas.

Additionally, destination information contributes to model stability. Experimental observations show that removing this field increases the likelihood of directional drift or deviation from the main channel in long-term predictions, consistent with the qualitative analysis discussed earlier for multimodal branching scenarios.

In summary, destination information, as a critical component of AIS static semantic attributes, has limited effect in short-term forecasting but significantly enhances the model’s ability to capture navigation intent in long-term predictions, effectively reducing trajectory deviation and endpoint error. These findings validate the necessity and rationality of the proposed multimodal fusion design.

\subsection{Ablation Study}\label{sec:4.4.7}
\subsubsection{Ablation Analysis of Key Modules}
\begin{table}[!t]
\centering
\caption{Ablation Study Results of M\textsuperscript{3}-Former\label{tab:ablation}}
\small
\begin{tabular}{@{}cccS[table-format=1.4]S[table-format=2.4]@{}}
\toprule
MoE & Multimodal & TurnLoss & {ADE$\downarrow$} & {FDE$\downarrow$} \\
\midrule
$\checkmark$ & $\checkmark$ & $\checkmark$ & \textbf{6.4727} & \textbf{13.6142} \\
$\checkmark$ & $\checkmark$ &             & 6.5665 & 13.7796 \\
             & $\checkmark$ &             & 6.6213 & 13.9759 \\
             & $\checkmark$ & $\checkmark$ & 6.6371 & 13.9431 \\
             &             & $\checkmark$ & 6.7146 & 14.1328 \\
             &             &             & 6.7732 & 14.3445 \\
$\checkmark$ &             &             & 6.9037 & 14.6456 \\
$\checkmark$ &             & $\checkmark$ & 6.9768 & 14.9049 \\
\bottomrule
\end{tabular}
\end{table}

To systematically assess the contribution of each module to the overall model performance, we conducted ablation experiments under identical training configurations by selectively removing or combining three key components: Multimodal semantic features (\textbf{Multimodal}), the dual-granularity Mixture-of-Experts structure (\textbf{MoE}), and the Steering-Weighted Loss function (\textbf{TurnLoss}). The results are summarized in Table~\ref{tab:ablation}, lower values indicate better prediction performance.

Overall, the complete model (\textbf{MoE + Multimodal + TurnLoss}) achieves the best performance, with ADE and FDE of 6.4727 and 13.6142, respectively. This indicates that the three proposed modules exhibit complementary effects when used jointly, significantly enhancing long-term trajectory prediction. Specifically, multimodal semantic features provide high-level navigation intent priors, the MoE structure enhances the model’s capability to capture complex dynamic patterns, and the turn-weighted loss emphasizes critical heading changes.

Regarding the effect of multimodal feature fusion, removing the \textbf{Multimodal} module (e.g., MoE + TurnLoss or MoE only) leads to notable performance degradation. For instance, with MoE and TurnLoss retained, ADE increases from 6.4727 to 6.9768, and FDE rises from 13.6142 to 14.9049. This indicates that relying solely on motion dynamics is insufficient to capture long-term navigation intent, whereas incorporating semantic information extracted via large language models (e.g., ship type, destination) effectively reduces uncertainty in long-term forecasts, thereby improving trajectory generation stability.

Concerning the MoE module, under identical multimodal input conditions, its inclusion further improves prediction accuracy. For example, in the \textbf{Multimodal + TurnLoss} configuration, removing MoE increases ADE from 6.4727 to 6.6371 and FDE from 13.6142 to 13.9431. This indicates that the dual-granularity MoE structure effectively captures variations across different navigation patterns, with expert routing mechanisms separately modeling local dynamic changes and global navigation strategies, thereby enhancing the model’s ability to represent complex trajectory distributions.

Finally, the impact of the turn-weighted loss function is evident in scenarios with complex heading changes. For example, under the \textbf{MoE + Multimodal} configuration, incorporating \textbf{TurnLoss} reduces ADE from 6.5665 to 6.4727. Given that straight-line motion dominates AIS trajectories, conventional loss functions tend to underemphasize turning behaviors. The turn-weighted mechanism addresses this imbalance by increasing the training weight for critical turning points, allowing the model to more accurately reconstruct trajectories in scenarios such as channel changes or port maneuvers.

In summary, the ablation results validate that all three core modules contribute positively to long-term vessel trajectory prediction. Multimodal semantic features provide global navigation intent priors, the MoE structure enhances modeling of complex dynamic patterns, and the turn-weighted loss mitigates training bias caused by imbalanced data distributions. Their synergistic effect enables the model to achieve optimal performance on both ADE and FDE metrics.

\subsubsection{MoE Structure Granularity Analysis}\label{sec:4.4.7.1}

\begin{figure}[htbp]
\centering
\includegraphics[width=0.95\linewidth]{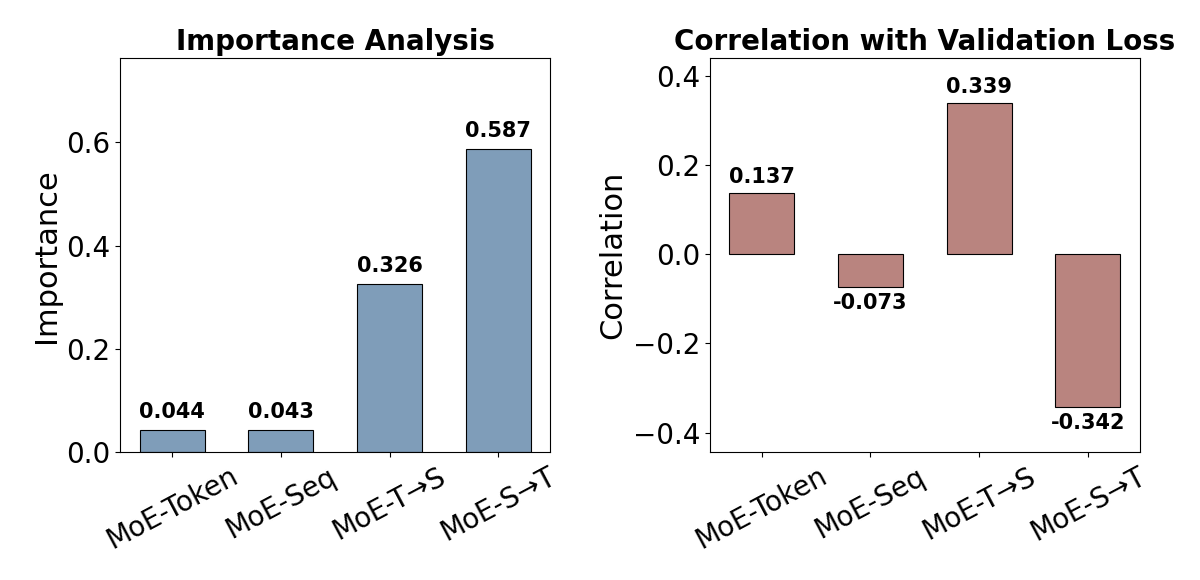}
\caption{Effect of Different MoE Granularity Designs on Prediction Performance.}
\label{fig:moe-granularity}
\end{figure}

\subsection{Effect of MoE Granularity on Model Performance}\label{sec:4.4.9}

To further evaluate the impact of different granularity designs within the Mixture-of-Experts (MoE) structure, we compared four MoE variants: token-level routing only (\textbf{MoE-Token}), sequence-level routing only (\textbf{MoE-Seq}), token-to-sequence routing (\textbf{MoE-T$\rightarrow$S}), and sequence-to-token routing (\textbf{MoE-S$\rightarrow$T}). The experimental results are summarized in Fig.~\ref{fig:moe-granularity}. Metrics are based on aggregated statistics over 20 runs. \textbf{Importance} measures the contribution of each structure to overall model performance, while \textbf{Correlation} reflects the relationship between the structure’s activation proportion and validation loss: positive correlation indicates an increase in prediction error, whereas negative correlation indicates a reduction in validation loss.

The results indicate that single-granularity MoEs (MoE-Token and MoE-Seq) contribute only modestly, with importance values of 0.044 and 0.043, and correlations near zero or slightly negative. This suggests that a single perspective is insufficient to fully capture complex navigation patterns: MoE-Token effectively models local dynamics but lacks global context, whereas MoE-Seq captures overall trends but ignores fine-grained variations.

In contrast, dual-granularity structures show substantial performance improvement. MoE-T$\rightarrow$S and MoE-S$\rightarrow$T have importance scores of 0.326 and 0.587, respectively, but exhibit different correlation characteristics. MoE-T$\rightarrow$S shows a positive correlation (0.339), indicating that despite its high importance, it can adversely affect validation loss. MoE-S$\rightarrow$T, on the other hand, exhibits a negative correlation (-0.342), suggesting that it is both highly important and effective at reducing validation error.

This observation can be explained by navigation principles: the sequence-to-token MoE-S$\rightarrow$T first captures global trajectory trends and subsequently models local dynamics at the token level, aligning with actual operational procedures—determine the overall route before applying local maneuvers. Conversely, the token-to-sequence MoE-T$\rightarrow$S is prone to propagate local noise, which can accumulate and interfere with global predictions, resulting in the “high importance but positive correlation” phenomenon.

Overall, the dual-granularity MoE effectively integrates local and global information. Among the variants, MoE-S$\rightarrow$T combines high importance with negative correlation, most closely reflecting the intrinsic patterns of vessel trajectory generation, providing empirical justification for its adoption in the final model architecture.

\section{Conclusion and Future Work}
\label{sec:conclusion_future}

This paper presented M\textsuperscript{3}-Former, a multimodal framework for long-term vessel trajectory prediction. By jointly modeling historical motion patterns and vessel static attributes, which include destination-related semantic information, the proposed method captures navigation intent and global route preferences. Built upon a discretization-based generative formulation, M\textsuperscript{3}-Former further introduces a dual-granularity mixture-of-experts architecture to model global semantics at the sequence level and local motion refinements at the token level. A steering-aware weighted loss is also adopted to better learn rare but important turning behaviors. Experimental results on real AIS data show that the proposed method consistently improves prediction accuracy and stability over strong baselines, especially in complex turning and route-branching scenarios.

Future work will focus on three directions. First, richer environmental factors, such as weather, sea state, tides, and traffic constraints, can be incorporated to improve prediction fidelity in complex maritime settings. Second, more efficient model designs are needed to support large-scale and real-time deployment. Third, transfer learning and domain adaptation should be explored to enhance generalization across different sea areas, vessel types, and traffic regimes. These extensions may further improve the practical value of multimodal trajectory prediction for maritime monitoring and decision support.

\section*{Acknowledgments}
This work is supported by the National Natural Science Foundation of China (NSFC) (Grant No. 52071312).



\bibliography{ref}

\bibliographystyle{IEEEtran}







\vfill

\end{document}